\documentclass{article}

\usepackage[a4paper, total={6in, 8in}]{geometry}

\usepackage{amsmath} 
\usepackage{ctable}
\usepackage{makecell}
\usepackage{subfigure}
\usepackage{caption}
\usepackage{xcolor}
\usepackage{soul}
\usepackage{hyperref}
\hypersetup{
    colorlinks=true,
    linkcolor=blue,
    filecolor=magenta,      
    urlcolor=cyan,
}

\title{\LARGE \bf
End-to-End Deep Learning Model for Cardiac Cycle Synchronization from Multi-View Angiographic Sequences*
}

\author{Raphaël Royer-Rivard$^{1}$, Fantin Girard$^{2}$, Nagib Dahdah$^{3}$ and Farida Cheriet$^{1}$
\renewcommand\footnotemark{}
\thanks{*Research supported by the TransMedTech Institute and the National Sciences and Engineering Research Council of Canada.}
\thanks{$^{1}$Department of Computer Engineering and Software Engineering of Polytechnique Montréal, Montréal, Canada
        {\tt\small raphael.royer.rivard@gmail.com}}%
\thanks{$^{2}$Research and Technology Unit at IDEMIA, Montréal, Canada}%
\thanks{$^{3}$CHU Sainte-Justine, Montréal, Canada}%
}

\begin{document}

\maketitle
\thispagestyle{empty}
\pagestyle{empty}

\begin{abstract}

Dynamic reconstructions (3D+T) of coronary arteries could give important perfusion details to clinicians. Temporal matching of the different views, which may not be acquired simultaneously, is a prerequisite for an accurate stereo-matching of the coronary segments. In this paper, we show how a neural network can be trained from angiographic sequences to synchronize different views during the cardiac cycle using raw x-ray angiography videos exclusively. First, we train a neural network model with angiographic sequences to extract features describing the progression of the cardiac cycle. Then, we compute the distance between the feature vectors of every frame from the first view with those from the second view to generate distance maps that display stripe patterns. Using pathfinding, we extract the best temporally coherent associations between each frame of both videos. Finally, we compare the synchronized frames of an evaluation set with the ECG signals to show an alignment with 96.04\% accuracy. 

\end{abstract}

\section{INTRODUCTION}

Coronary Angiography (CA), a conventional catheter-based 2D x-ray imaging technique, has been widely used over the years and is still considered as the gold standard for diagnosing and guiding coronary artery interventions. Since a single view of CA cannot show every detail of the whole coronary tree, multiple sequences are recorded from different angles, which cannot always be taken simultaneously.

A 3D reconstruction of the coronary tree from the angiographic sequences can help clinicians visualize morphological anomalies better than 2D images alone \cite{Garcia}. To obtain the most accurate 3D model from 2D images, synchronization between the sequences is mandatory. For this purpose, ECG based gating is common practice to help prevent artifacts caused by the cardiac movement \cite{Dattilo}. However, this method removes the temporal dimension of the angiographic sequences, which prevents the extraction of functional information \cite{Benovoy}. Furthermore, artifacts in the ECG signals can mislead automated algorithms that identify expected waveforms 
for cardiac phases identification.

A few recent works have tackled the cardiac phase synchronization task by relying on image processing. The authors of \cite{Panayiotou} proposed a statistical method for cardiorespiratory gating on fluoroscopic images, which was also later used in \cite{Toth}, but that method fails when the contrast agent is injected unevenly because of image intensity fluctuations. The work in \cite{Song} made use of multi-layer matching for motion tracking to determine the cardiac phase by measuring the cardiac motion velocity of local patches of coronary arteries. While this method was shown to work relatively well to synchronize angiographic sequences, the method completely relies on the accurate tracking of the vessel segments, which is not always possible.

Another recent work used a Deep Convolutional Neural Network (DCNN) to train a model to identify cardiac phases using a training dataset of several thousand coronary angiographies \cite{Ciusdel}. Unfortunately, that amount of data is not always available. To overcome this problem, transfer learning can be helpful. Some works have shown that transfer learning with finetuning made possible the task of video synchronization, although these approaches used unsupervised learning on non-cyclic videos \cite{Wieschollek,Dwibedi}.

In this work, we introduce a robust and fast method for synchronizing two angiographic sequences with respect to the phase in the cardiac cycle. 
The contributions are as follows: i) Training of the DCNN model is done in an end-to-end manner to output correspondences between video frames of different views from raw angiographic images. ii) The need for abundant angiographic data is avoided by making use of transfer learning. iii) A novel loss function, called the soft pair loss, is proposed for training a neural network from weakly annotated data to extract features describing cyclic video frames.

\section{METHOD}

\begin{figure*}[!tb]
    \centering
    \includegraphics[width=1.0\linewidth]{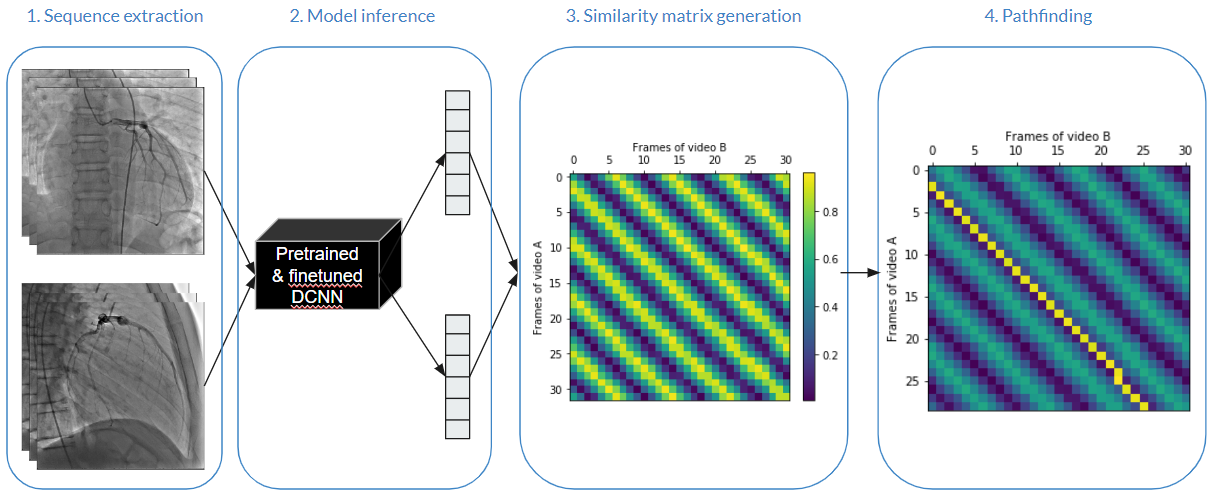}
    \caption{\textbf{Overall Processing Pipeline. 1. Sequence extraction}: Sequences of 3 consecutive frames from 2 videos in different viewing angles are given as inputs to the model. \textbf{2. Model inference}: Inputs are used for feedforward passes of a DCNN pretrained on ImageNet and finetuned on our data to output feature vectors (1 per sequence of 3 images). \textbf{3. Similarity matrix generation}: Every feature vector of video A is compared with every feature vector of video B to generate a similarity matrix. \textbf{4. Pathfinding}: After inverting the intensities of the matrix to convert the similarity values into costs for the pathfinding algorithm, the best path is identified. Each point of the path represents a synchronized frame pair between the two videos.}
    \label{figure_synchronization_pipeline}
\end{figure*}

The proposed method consists of four steps. First, the training images are selected automatically from angiographic sequences based on their global contrast and stacked in short sequences of 3 consecutive frames. Second, the short sequences are used as input to a pretrained and finetuned DCNN to extract their features. Third, the features extracted from the short sequences from two different views are compared using a cosine similarity measure to create a similarity matrix. Fourth, pathfinding is used on the similarity matrix to identify the best temporally coherent synchronization between the two videos. The described pipeline is shown in Fig. \ref{figure_synchronization_pipeline}.

\subsection{Sequence extraction}

We utilize only the sections of the original CA videos in which the contrast agent is fully visible in the coronary arteries. 
To this end, we keep only the frames between the point at which the global pixel intensity gradient rises above the average of the video and the point at which it falls back under the average. The images are downsized from 1024x1024 pixels to 224x224 or 248x248 pixels to comply with the expected input size of the selected model. Additionally, images are stacked in sets of 3 consecutive frames to create short sequences that are used as inputs to the model. This is done in a sliding window fashion, so most of the images are present in different positions in 3 sequences.

\subsection{Model inference}

To extract meaningful features representing the coronary arteries' movement, a network structure capable of locally parsing multiple consecutive angiographic images is required. Convolutional neural networks are well suited for this kind of task. Among the available pretrained models, we tested several of the best performing ones in the ImageNet Large Scale Visual Recognition Challenge \cite{Russakovsky}, notably ResNet-50 \cite{He}, MobileNet-v2 \cite{Sandler} and EfficientNet \cite{Tan}. To include the temporal aspect in our input data, we use the short sequences of 3 consecutive monochrome images. This approach also allows us to keep the first convolutional layer of the network as is, since it was designed to take as input three-channel (RGB) images.

\subsection{Similarity matrix generation}

The finetuned neural network is used to generate feature vectors from every combination of short sequences for two distinct videos. From these feature vectors, a similarity matrix is generated by computing the cosine similarity between every feature vector of video A and those of video B. Note that this matrix displays a diagonal stripe pattern as seen in Fig. \ref{figure_synchronization_pipeline}.

\subsection{Pathfinding}

To find the optimal synchronization between the two videos, the diagonal lines of maximal similarity 
in the similarity matrix must be identified. For this purpose, Dijkstra’s algorithm is run multiple times on the matrix with different starting locations. At the first iteration, the starting points are located at every 5 pixels of the first row and at every 5 pixels of the first column. The pathfinding is allowed to search only in the downward, rightward and diagonal directions. A maximum of 3 consecutive steps in the same horizontal or vertical direction is enforced to prevent the pathfinding from diverging from a diagonal line, which can otherwise happen in low contrast portions of the videos. The pathfinding stops when it reaches the last row or last column of the matrix. Every distinct ending point becomes a starting point for the second iteration, where the pathfinding is run in the opposite direction. Every path of a sufficient length 
is considered for the selection of the best path. Among the longest paths found, the straightest one is identified as the solution and is used to evaluate the performance of the model.

\subsection{Training strategy}

\subsubsection{ECG}

To generate the ground truth used by the training process to improve the pretrained neural network, an algorithm that parses ECG signals recorded with the angiographic sequences was developed. This algorithm detects the peaks of the R-wave parts of the signal by locating the local maxima of the smoothed and normalized signals using gaussian kernels and gradients. The detected peaks of the R-waves were manually inspected to ensure their quality and the algorithm failed to identify the peaks in only a few cases where the signal suffered from a measurement problem due to the sensor’s sensitivity range. These errors were not manually corrected and were considered as tolerable noise for the training process.

\subsubsection{Ground truth}

The R-waves peaks extracted from the ECG were used to generate a ground truth matrix where every element $(t,t')$ represents the synchronization level ranging from 0 to 1 between frame $t$ of video A and frame $t'$ of video B. A synchronization level of 1 means that frames $t$ and $t'$ are located at exactly the same phase in their respective cardiac cycles, while 0 means there is half a cardiac cycle in distance between their cardiac phases. The cardiac phase is computed from the cardiac cycle progression using the R-wave peaks. More precisely, since a cardiac cycle is comprised of all video frames between two consecutive R-wave peaks, the cardiac phase is the proportion of total frames between frame $t$ (resp. $t'$) and the peak preceding it. 
This method of creating the ground truth matrix from the ECG signals of both videos allows for a good temporal correspondence at the end of diastole phase, even when the difference between the heart rates is high.

\subsubsection{Soft pair loss function}

During DCNN training, for each mini-batch, $N$ random frames are sampled in a random video, where $N$ is the size of the mini-batch. For every sampled frame, the 3-channel input image is created by stacking that frame with the two previous frames of the video. Given $i$ and $j$, two sampled input images from the mini-batch, we generate the feature vectors $\mathbf{a}$ and $\mathbf{b}$ from forward passes of the neural network. The cosine similarity is computed as the dot product of these feature vectors divided by the product of their norms:

\begin{equation}
    \text{cos}(\theta)_{i,j} = \frac{\mathbf{a} \cdot \mathbf{b}}{\Vert \mathbf{a} \Vert\Vert \mathbf{b} \Vert}
\end{equation}

This cosine similarity, ranging from -1 to 1, indicates the similarity between the directions of the two feature vectors. The loss function used to finetune the network is simply equal to the average squared distance between the normalized cosine similarity of every feature vector pair and its respective ground truth similarity $y_{i,j}$.

\begin{equation}
    \text{loss} = \sum_{i=1}^{N}\sum_{j=1}^{N}\frac{(0.5(\text{cos}(\theta)_{i,j}+1) - y_{i,j})^2}{N^2}
\end{equation}

Standard pair losses and triplet losses used by other siamese networks, i.e. networks that use the same weights for processing two different inputs to produce comparable output vectors,
generate only a limited amount of valid feature vector combinations per mini-batch due to the fact that every training example needs a positive and negative example. By contrast, our loss function, which we call the soft pair loss, generates a richer learning signal from the sampled feature vectors since every vector can be matched with all the others.

\section{RESULTS \& DISCUSSION}

The dataset used to train the neural network was composed of 271 free breathing angiographic sequences of the left or right coronary artery tree from 39 distinct pediatric patients at CHU Sainte-Justine. The videos had 117.8 frames on average and were recorded over an average of 8.8 cardiac cycles from a wide variety of angles at 15 or 30 frames per second using a Toshiba Infinix-CFI BP system. After extracting the video sections with visible contrast agent, the resulting dataset had an average of 53.6 frames per video and 4.2 cardiac cycles. 
39 sequences from 5 patients were used to create the validation set and 30 sequences from 5 other patients composed the testing set. The rest of the data was used for training.

\subsection{Score calculation}

For every video combination of each patient in the test set, 
the similarity matrix is generated and the best synchronization path is found. The score of the path is then defined as the average value of its points mapped onto the corresponding ground truth matrix. 
To normalize the score, it is then divided by the score of the path found directly on the ground truth matrix, i.e. the maximum possible score for that specific synchronization.

\subsection{Training hyperparameters}

To optimize the model parameters, a random hyperparameter search was carried out. The explored parameters included several standard ones, but some were related to the specific task of cardiac cycle synchronization, namely the maximum number of cardiac cycles between two frames of a training pair and the use of different videos to create training pairs (when disabled, training pairs were created only with frames from the same video). The hyperparameters of the best model for each architecture and the resulting overall scores are listed in Table \ref{table_hyperparameter_comparison}. The whole finetuning process of 20 epochs took only 2 hours on a desktop PC with an Nvidia GeForce RTX 2070 graphics card with 8 GB of RAM.

\begin{table}[!tb]
    \caption{Hyperparameter comparison of our best models.}
    \label{table_hyperparameter_comparison}
    \begin{center}
        \begin{tabular}{cccccccc}
            \specialrule{.2em}{.1em}{.1em}
            Model Type & FC & BS & DR & MC & IVP & DA & Score\\
            \specialrule{.2em}{.1em}{.1em}
            ResNet-50 & 64 & 32 & 0.183 & 0 & false & true & 0.9549\\
            MobileNet & 128 & 32 & 0 & 2.02 & false & false & 0.9558\\
            EfficientNet-b0 & 8 & 16 & 0 & 0 & true & true & 0.9581\\
            EfficientNet-b1 & 4 & 12 & 0 & 0 & false & false & \textbf{0.9604}\\
            \specialrule{.2em}{.1em}{.1em}
        \end{tabular}
    \end{center}
    \caption*{\footnotesize\textbf{FC}: size of Fully Connected layer, \textbf{BS}: Batch Size, \textbf{DR}: Dropout Rate, \textbf{MC}: Max Cycles, \textbf{IVP}: Inter Video Pairs, \textbf{DA}: Data Augmentation.}
\end{table}

\subsection{Analysis}

\begin{figure}[!tb]
    \centering
    \includegraphics[width=1.0\linewidth]{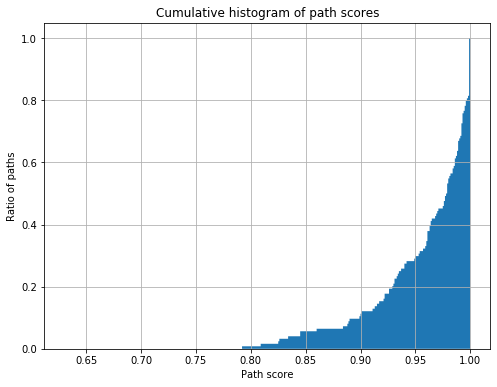}
    \caption{\textbf{Cumulative histogram of normalized path scores} for the video pairs in the test set.}
    \label{figure_cumulative_histogram}
\end{figure}

While the average score of the best model is 0.9604, the median is significantly higher at 0.979. By extracting the path scores for all the video pairs in the test set, as shown in the cumulative histogram of Fig. \ref{figure_cumulative_histogram}, we see that some of the scores are fairly low. Indeed, around 5\% of the synchronization paths have a score under 0.85 based on the ECG ground truth. Among the 124 video pairs of the testing set, 8 of them are sequences acquired simultaneously with a biplane C-arm. As we could expect, these pairs have a higher score of 0.9857, while the rest have a lower than average score of 0.9586.

\begin{figure}[!tb]
    \centering
    \includegraphics[width=1.0\linewidth]{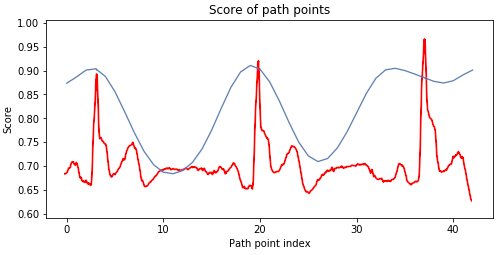}
    \caption{\textbf{Example of low-scoring path.} The blue curve represents the unnormalized value of each point in the path; the red curve shows the recorded ECG signal. The point values tend to be lower in-between R-wave peaks.}
    \label{figure_selected_path}
\end{figure}

Fig. \ref{figure_selected_path} shows the individual path point values for an example of low-scoring path. Here,  
we observe a tendency for path points to get higher values (in the ground truth matrix) near the synchronized R-wave peaks (end of diastole) than in-between (end of systole). When qualitatively evaluating the synchronization accuracy from the video frame pairs, visual examination revealed that the synchronization based on the selected path appeared more accurate than when based on the ECG ground truth. 
Indeed, the latter seemed to be off by a few frames at the end of systole phases, while the former looked relatively well synchronized throughout the whole sequence. This suggests that our method can outperform synchronization based solely on the R-wave peaks of the ECG signals.

Additionally, the pathfinding algorithm tends to deviate from the ground truth at the boundaries of the similarity matrix, 
which can be explained by the fact that imperfections in feature extraction can be mitigated by the rest of the path at other locations but not at the extremities. A simple solution would be to trim the first and last few frames of the synchronized pairs. Indeed, removing the first and last points of every path slightly increases the average score from 0.9604 to 0.9607.

\section{CONCLUSION}

In this paper, we demonstrate that cardiac cycle synchronization of multi-view angiographic sequences can be performed with better accuracy using our novel method than using ECG signal gating on R-wave peaks. 
Our deep learning method needs only few training examples, curtails problems related to ECG signal quality and availability by focusing on image sequence analysis instead, and provides a basis for new approaches to 3D reconstruction of the coronary tree from angiography.

\end{document}